\documentclass[letterpaper, 10 pt, conference]{ieeeconf}  

\usepackage{mathtools}

\IEEEoverridecommandlockouts                              

\usepackage{color, colortbl}
\usepackage{array}
\usepackage{fancyhdr}
\title{\LARGE \bf
Situation Assessment for Planning Lane Changes: Combining Recurrent Models and Prediction
}

\author{Oliver Scheel$^{1, 2}$, Loren Schwarz$^1$, Nassir Navab$^2$, Federico Tombari$^2$
\thanks{$^1$Oliver Scheel and Loren Schwarz are with the BMW Group, 80788 M\"unchen, Germany, {\tt\small \{oliver.scheel, loren.schwarz\}@bmw.de}}%
\thanks{$^2$Federico Tombari and Nassir Navab are with the Faculty of Computer Science,
        Technische Universit\"at M\"unchen, 85748 Garching bei M\"unchen, Germany, {\tt\small \{tombari, navab\}@in.tum.de}}%
        }

\begin{document}

\setlength{\textfloatsep}{5pt plus 1.0pt minus 2.3pt}

\maketitle
\thispagestyle{fancy}
\pagestyle{empty}

\begin{abstract}
One of the greatest challenges towards fully autonomous cars is the understanding of complex and dynamic scenes. Such understanding is needed for planning of maneuvers, especially those that are particularly frequent such as lane changes.
While in recent years advanced driver-assistance systems have made driving safer and more comfortable, these have mostly focused on car following scenarios, and less on maneuvers involving lane changes. In this work we propose a situation assessment algorithm for classifying driving situations with respect to their suitability for lane changing. For this, we propose a deep learning architecture based on a Bidirectional Recurrent Neural Network, which uses Long Short-Term Memory units, and integrates a prediction component in the form of the Intelligent Driver Model. We prove the feasibility of our algorithm on the publicly available NGSIM datasets, where we outperform existing methods.
\end{abstract}

\section{INTRODUCTION}
Nowadays, most modern cars come with at least some advanced driver-assistance systems (ADAS). Systems like an automatic cruise control or a lane keeping assistant are already able to partially take over control of the car and safely steer it along a defined lane. While these problems have been addressed extensively in scientific literature \cite{ZHENG201416}, research about lateral control involving lane changes has not been studied as intensively up to now \cite{8005678}. For the challenge of driving fully autonomously, this naturally has to be addressed as well. Additionally, there is great potential and need for driver assistance systems supporting the driver in executing lane changes. Over 90\% of occurring accidents are attributed to human errors \cite{trafficsafetly}, of all accidents around 18\% happen during the execution of a lane change \cite{tsa}.

The term driving strategy describes planning for autonomous vehicles on different hierarchical levels, from map-based global mission planning to tactical planning, which takes into account driving lanes, other vehicles and obstacles. From a machine learning perspective one way to approach this problem is the use of supervised learning. Possible is for example behavioral cloning, in which a system learns from the demonstration of an expert \cite{NIPS2007_3293}. However, it is well known that small inaccuracies and errors aggregate over time \cite{pmlr-v9-ross10a}, further this method tends to overfit specific expert behavior and lacks the possibility of exploration.
Another possibility is the application of reinforcement learning, which lately has led to great success in different topics \cite{DBLP:journals/corr/MnihKSGAWR13}. This method though often depends on the availability and realism of a simulator, and discards the immense amounts of real data collected by car manufacturers.
We believe a combination of both paradigms will be necessary for creating data-driven algorithms for autonomous cars.

\begin{figure}[!t]
      \centering
      \includegraphics[scale=0.27]{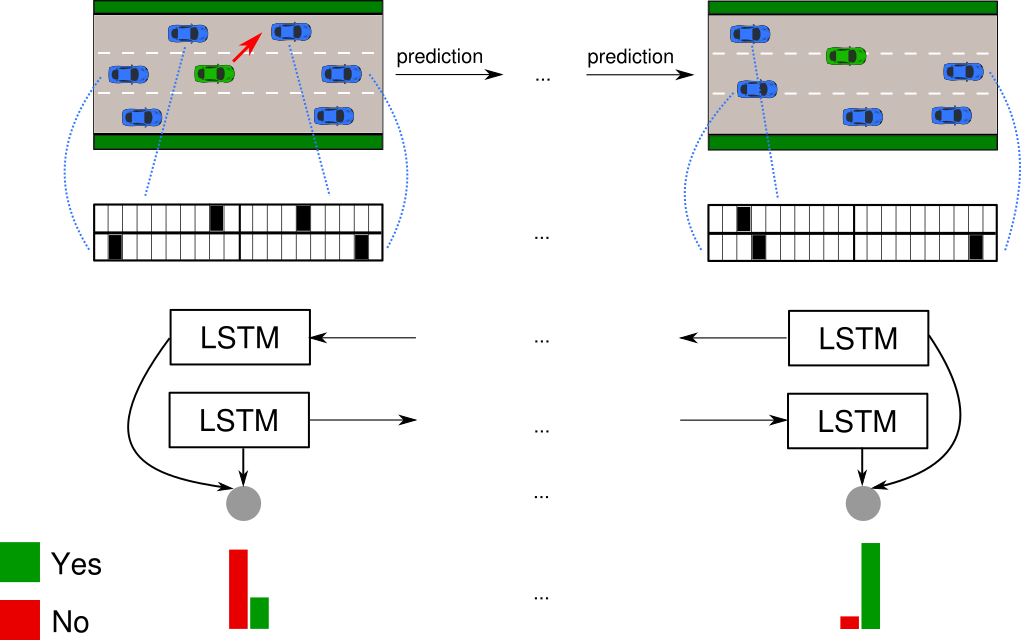}
      \caption{Visualization of our system assessing a situation. For a timestep $t$ the LSTM network already has a compressed understanding of the past via its internal state, with the IDM several future positions are predicted for enabling the use of a Bidirectional LSTM. The situations are converted to the internal grid representation and fed to the network, the output represents the suitability of a situation for a lane change to the target lane.}
      \label{figureteaser}
   \end{figure} 
   
An autonomous system is commonly designed in a layered architecture: A perception layer perceives the environment through different sensors (1), the results are fused in a fusion layer (2). Based on this a situation interpretation is done (3). This is followed by a planning (4) and control layer (5). This work is situated around layer 3, supporting the planning algorithm, thus circumventing the issues mentioned above arising by using machine learning for planning. Our algorithm is able to answer the question whether a given situation is safe and suited for executing a lane change. It is trained in a supervised fashion on actual driving data, and is easily extensible to other discrete driving decisions by simply exchanging the training data. A possible application not only is the integration into fully autonomous cars, but also some ADAS to support the driver in lane change decisions.
We compare our proposed method to existing ones and show its superiority.
As finding useful data labels is a challenge, we additionally give details about existing labeling methods and propose a new automatic labeling scheme.  

Recently more interest has sparked in lateral planning \cite{7795631, BALAL201647, 7313255, 7576883}, possibly due to the announcement of several car manufacturers to produce autonomous cars within the next years \cite{bmw21, tesla21}. One big field is reinforcement learning, in which deep networks have been applied to decision making for lane changes \cite{A1}. This though somewhat differs from our task due to our focus on situation assessment without planning using supervised learning. 
More related to our field mostly either mathematical models \cite{nagel}, rule-based systems \cite{BALAL201647} or ``classical'' machine learning methods like Support Vector Machines (SVMs) or decision trees \cite{7795631, 8005678} are used.
In this paper, we propose the first supervised deep learning approach on non-visual data, namely a Recurrent Neural Network (RNN) with Long Short-Term Memory (LSTM) units. We further extend this to a Bidirectional RNN. As Bidirectional RNNs either need to know the future of an observed signal or introduce a significant time lag, we integrate a prediction component, for which we use the Intelligent Driver Model (IDM). Fig. \ref{figureteaser} shows our system evaluating a situation with respect to its suitability for a lane change to the left.
We believe that deep learning greatly helps in the understanding of complex scenes and that the recurrent structure is much better suited for understanding the temporal component of this problem, compared to static frame-by-frame approaches. The evaluation of our method is done on the publicly available NGSIM datasets.
The key contributions of our work are:
\begin{itemize} 
\item We propose a method for assessing situations with respect to their suitability for lane changes, which is easily extensible to other driving decisions.
\item Our proposed method consists of a Bidirectional RNN and is the first recurrent approach for this problem. It uses the IDM for obtaining an explicit representation of the beliefs over future states.
\item We propose a novel labeling method for this kind of problem and compare it to the one used in existing works.
\end{itemize}

\section{RELATED WORK}
In recent years first partially feasible algorithms for fully self-driving cars, capable of executing lane changes, have been developed and tested on the road. Mostly these were rule-based systems. 
A probabilistic framework for decision making was developed by Ardelt et al. and tested during autonomous drives on highways \cite{6200871}. 
Ulbrich and Maurer \cite{7313255} used Dynamic Bayesian Networks (DBNs) to evaluate situations to answer the question whether they are feasible and beneficial with respect to lane changes.
Men\'{e}ndez-Romero et al. introduced a multi-level planning framework which makes use of semantic and numeric reasoning \cite{7995915}.
Mukadam et al. used deep reinforcement learning for making lane change decisions \cite{A1}. 

In existing literature a lane change maneuver is often categorized into one of three categories: Mandatory lane changes are lane changes forced upon the driver, e.g. due to traffic conditions like an ending lane.
Discretionary lane changes are performed to improve driving conditions and anticipatory lane changes are done preemptively to avoid, for example, future traffic congestions. Many lane change decision making models partition the process of a lane change in several phases, which start with the decision to change lanes and end with the acceptance of a suitable gap \cite{7795631, BALAL201647}. Assessing gaps thus is a crucial part, and usually a binary classification problem.

Several approaches have been proposed for modeling and predicting human behavior. Dou et al. \cite{7576883} considered mandatory lane change events at lane drops and predicted driver merging behavior with SVMs and simple Feedforward Neural Networks. 
Different algorithms were compared by Motamedidehkordi et al. \cite{8005678} for solving the same problem of learning human driving behavior and predicting lane change decisions. The tested algorithms included, amongst others, SVMs and decision trees. 

Another important and related problem is the prediction of future driving situations.
Recently LSTMs have been adopted by the autonomous driving community for this, leading to good results in predicting trajectories and future driving maneuvers, outperforming non-recurrent architectures \cite{7565491}.

Most related and comparable to our work are binary classification problems assessing the suitability of a situation for a lane change, as the previously mentioned work from Ulbrich \cite{7313255}.
Nie et al. \cite{7795631} described a gap assessment model for discretionary lane changes using SVMs. For evaluation purposes a selected subset of the NGSIM US Highway 101 (US 101) dataset was used. Their approach outperformed previous ones, correctly classifying 97.48\% of occurring lane changes. 
Balal et al. \cite{BALAL201647} addressed the problem using fuzzy logic rules, with the aim of supporting the driver in lane change decisions. The NGSIM Interstate 80 (I-80) and the US 101 dataset were used while considering only certain lanes and time periods. Correct recommendations were given in 90.50\% of the cases. Jeong et al. used Convolutional Neural Networks (CNNs) to classify neighbored lanes as free or blocked based on camera data \cite{7995938}.
\section{PROBLEM FORMULATION}
Our goal is to classify situations into the categories suited and not suited for a lane change (one can also speak of safe and unsafe). Once a target lane has been identified, for each timestep such a classification has to be done. 

\begin{figure}[!t]
      \centering
      \includegraphics[scale=0.3]{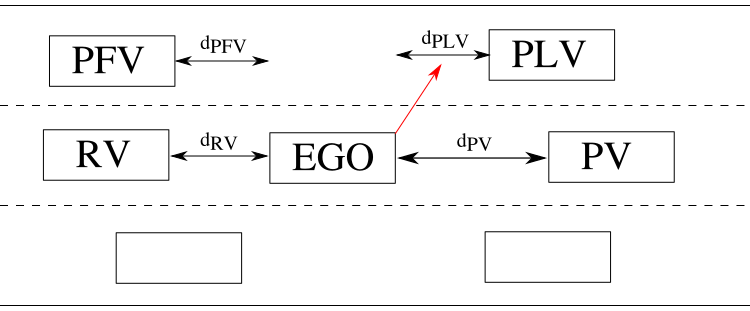}
      \caption{Cars involved in a lane change decision.}
      \label{figurestreet}
   \end{figure}

We adopt the notation of Nie et al. \cite{7795631} concerning involved cars and input features, see Fig. \ref{figurestreet}. 
The ego car, for which we want to conduct the situation assessment, is marked by EGO. Its preceding and following car on the same lane are denoted by PV (preceding vehicle) and RV (rear vehicle), respectively. The preceding and following cars on the target lane are called putative leading vehicle (PLV) and putative following vehicle (PFV). Let $x_c$ be the longitudinal coordinate of vehicle $c$, $c \in \{EGO, PV, RV, PFV, PLV \}$, on the rectified projection of its current driving lane and $v'_c$ its velocity. Then the distance from the ego car to another vehicle $c$  is denoted by $d_c = \vert x_{EGO} - x_{c}\vert$, the relative velocity by $v_c = v'_{EGO} - v'_c$ and the temporal distance by $t_c = \frac{d_c}{v_c}$. 

Let $f^t$ be a data frame at timestep $t$. Our algorithm assigns a classification $o^t$ to the frame, where $o^t \in \{1, 0\}$ is a label expressing that in timestep $t$ a lane change is possible (safe) or not. Denote the ground truth with $l^t \in \{1, 0\}$. As evaluation metric we use the average accuracy $acc = \frac{acc_p + acc_n}{2}$, where $acc_p$ and $acc_n$ are the fractions of correctly classified frames with $l^t = 1$ and $l^t = 0$, respectively.

\section{A RECURRENT MODEL FOR ASSESSING SITUATIONS}
\label{sectionrnn}
This section covers our main contribution, namely an LSTM network for assessing driving situations and its extension, a Bidirectional LSTM.
Bidirectional RNNs use information from future time steps. For training this is feasible, but not for an online implementation. Thus we include a prediction component in the network. This is essentially a black-box algorithm and can be any valid trajectory prediction algorithm. Here we use the IDM.
We train the models in a sequence-to-sequence fashion, meaning for each timestep $t$ we generate an output $o^t \in \{0, 1\}$, indicating the suitability of the situation for a lane change. As input features only  $d_{PV}$, $d_{RV}$, $d_{PFV}$ are $d_{PLV}$ are passed to the network, in contrast to previous models \cite{7795631, BALAL201647}, as we expect the networks to learn an understanding of velocity (and more) out of the positions.
\subsection{Intelligent Driver Model}
To obtain a practically feasible Bidirectional RNN we need a component which is able to predict vehicle positions for the near future. 
We use the Intelligent Driver Model (IDM) \cite{2000PhRvE..62.1805T}.
It is a car-following model, which describes a rational driver, who tries to keep a safe distance to the preceding car in dense traffic, but accelerates up to a certain desired velocity when this is feasible.
Denote with $X$ the vehicle in question and with $Y$ the vehicle directly in front of it. Let $x_c$ be the position of vehicle $c$ at time $t$, $v'_c$ its speed and $l_c$ its length, $c \in \{X, Y\}$. Define $s_X = x_Y - x_X - l_Y$ and $v_X = v'_X - v'_Y$ Then the model is described by the following system of differential equations:
\begin{equation}
\begin{split}
\dot{x_{X}} & = \frac{dx_{X}}{dt} = v'_{X} \\
\dot{v}'_{X} & = \frac{dv'_{X}}{dt} = a(1 - (\frac{v'_{X}}{v_0})^{\delta} - (\frac{s^*(v'_{X}, v_{X})}{s_{X}})^2)  \\
s^*(v'_{X}, v_{X}) & = s_0 + v'_{X}T + \frac{v'_{X} v_{X}}{2 \sqrt{ab}}
\end{split}
\end{equation}
$v_0$, $s_0$, $T$, $a$, $b$ and $\delta$ are freely choosable model parameters with the following meaning: $v_0$ is the desired velocity, 
$s_0$ a minimum spacing,
$T$ is the desired time headway.
$a$ describes the maximal vehicle acceleration,
$b$ the comfortable braking behavior. 
$\delta$ is some exponent.
Best prediction results could of course be obtained by considering all cars in a scene, however this is not feasible, as for an online implementation we can only examine cars seen by the sensors of the ego car. With modern Radar sensors though it is definitely possible to spot and detect the surrounding vehicles PV, RV, PFV, PLV and even the cars preceding or trailing these (if these cars are within a certain distance of the ego car, but if not their influence in the IDM is negligible anyways). 
Thus for each needed future timestep we use the IDM to predict the new velocity and position of the vehicles PV, RV, PFV, PLV and EGO. For EGO, RV and PFV our prediction will be more accurate, as their preceding vehicles PV and PLV are part of the prediction model and will be reevaluated in each step. For the vehicles PV and PLV we simply observe their preceding cars at the current moment, and from then on assume a constant velocity over the whole prediction period.
\subsection{Long Short-Term Memory Network}
In this subsection we introduce the non-predictive LSTM model, see \cite{gers1999learning} for a definition of LSTMs.
Experiments showed best performance when converting the inputs into an occupancy grid and embedding this before feeding it to the LSTM. The grid is partitioned into four parts - same lane before EGO; same lane behind EGO; target lane before EGO; target lane behind EGO. Each part describes 100 meters of road and is discretized into boxes representing 10 meters each. Only the vehicles PV, RV, PLV and PFV are considered and represented by a 1 in the corresponding vector, the remaining entries are filled with 0s. This way the embedding process resembles the embedding of discrete words, amongst others used in machine translation \cite{DBLP:journals/corr/MikolovSCCD13}, which inspired our decision.
The output of the LSTM is transformed with a softmax function into a probability distribution, from which the maximum is taken as the final classification of the network. The full network is described by the following equations, see Fig. \ref{figurelstm} for a visualization:
\begin{figure}[!t]
      \centering
      \includegraphics[scale=0.20]{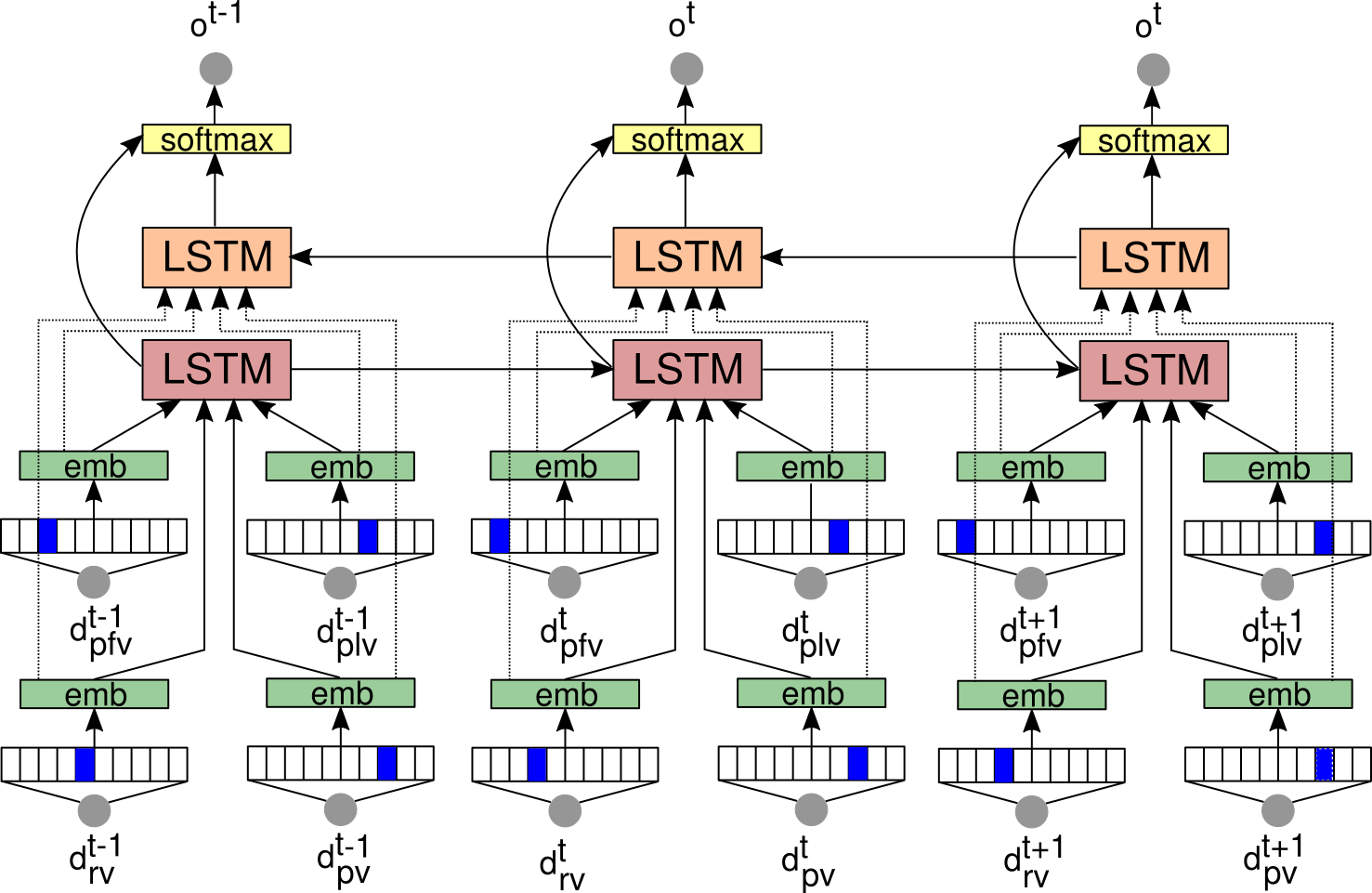}
      \caption{Visualization of the LSTM network. The observed inputs $d_x$ are converted into the grid representation $g_x$, embedded and fed to the LSTM units. The output $o^t$ is obtained by applying a softmax function to the output of the LSTM units. The standard LSTM network consists only of the red colored LSTM cells, for the Bidirectional LSTM network also the orange ones are used. This additional layer processes information in a reversed temporal order.}
      \label{figurelstm}
   \end{figure} 
\begin{equation}
\begin{split}
e^t & = emb(W_{emb}, b_{emb}, [g^t_{pv}; g^t_{rv}; g^t_{pfv}; g^t_{plv}] )\\
(h^t, c^t) & = LSTM(e^t, h^{t-1}, c^{t-1})\\
o^t & = softmax(W_o h^t + b_o)
\end{split}
\end{equation}
Here $g^t_x$ describes the one-hot vector for vehicle $x$ obtained by the discretization of $d^t_x$ described above, and $emb(W, b, [x_1, x_2, \ldots]) = [W x_1 + b; W x_2 + b; \ldots]$.  The function $LSTM$ denotes all calculations done in a single LSTM unit, returned are the unit's output representation $h^t$ and its internal state $c^t$. 
\subsection{Bidirectional Long Short-Term Memory Network}
Bidirectional LSTMs represent an extension of standard LSTM networks by adding another LSTM layer, which processes information backwards. The network then does not only have access to information from the past, but also from the future. In each step, the output of these otherwise separated layers is combined and further processed to obtain the final output.

For training real future trajectories are used, for testing predicted ones with the help of the IDM. This anticipation helps the algorithm assess traffic situations, as situation assessment is never a static problem. A prediction, which happened implicitly in, for example, the previously mentioned LSTM or an SVM model, now is done explicitly. Consider a situation in which a fast car approaches the ego vehicle in the target lane from behind, but starts braking strongly. At first a lane change might not seem safe, but it might well be after the braking maneuver. A conventional situation assessment algorithm might decide likewise, due to the detected braking, implicitly calculating future vehicle positions and resulting gaps. Our proposed bidirectional LSTM does this explicitly by querying a prediction component about future states. 
We concatenate the outputs of both LSTM units before feeding them to the softmax layer.
The full modified model looks like this, compare Fig. \ref{figurelstm}:
\begin{equation}
\begin{split}
e^t & = emb(W_{emb}, b_{emb}, [g^t_{pv}; g^t_{rv}; g^t_{pfv}; g^t_{plv}] )\\
(h^t_F, c^t_F) & = LSTM(e^t, h^{t-1}, c^{t-1})\\
(h^t_B, c^t_B) & = LSTM(e^t, h^{t+1}, c^{t+1})\\
o^t & = softmax(W_o [h^t_F; h^t_B] + b_o)
\end{split}
\end{equation}
Let $C = \{PV, RV, PLV, PFV\}$.
For training in each iteration we feed sequences of length $T_F$ to the network, in which $g^t_x$ is derived from real trajectories for each $t \in \{1, \ldots, T_F\}$, $x \in C$. While doing so we reset the internal state of the backward LSTM, $c_B$, after each $T_B$ steps. This way at each timestep the network is limited to seeing at most $T_B$ future steps, which is the prediction horizon of our prediction component. During testing, in each iteration we feed sequences of length $T_B$ to the network, repeating the following for each time step $t \in \{1, \ldots, T_B\}$: $g^t_x$, $x \in C$ is derived from real vehicle positions, all $g_{t'}^x$, $t' \in \{t, \ldots, T_B\}$, $x \in C$ are derived from predicted trajectories. The output $o^t$ is saved and appended to the list of final outputs.
We use $T_F = T_B$ = 10 seconds, note though that arbitrary values are possible, especially with $T_F > T_B$. Greater values for $T_F$ and $T_B$ might increase the performance of the algorithm, it is recommended to set $T_B$ as large as a reliable prediction from the prediction component can be guaranteed.

\section{DATA DESCRIPTION AND DATA LABELING}
Since our solution relies on supervised learning, data labeling is necessary. For each data frame $f^t$ we want to annotate it with a label $l^t \in \{0, 1\}$, indicating whether $f^t$ is suited for a lane change or not.
In this section we first briefly introduce the used NGSIM datasets. Then we describe a labeling approach used in previous works and briefly discuss its drawbacks. Eventually we propose an improvement of this principle and additionally introduce a new labeling method.

\subsection{NGSIM Dataset}
\begin{figure}[!t]
      \centering
      \includegraphics[scale=0.3]{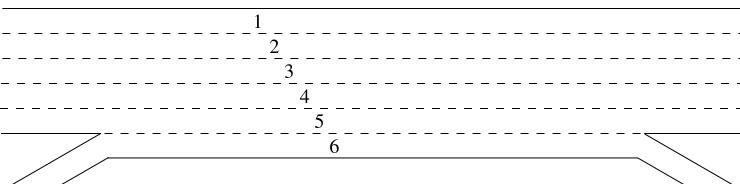}
      \caption{Area of study in US 101 dataset (size proportions not realistic). The I-80 dataset looks similar, except it is missing the off-ramp.}
      \label{figureus101}
   \end{figure}

The Next Generation Simulation (NGSIM) project contains several publicly available traffic data sets \cite{ngsim}. Here we use the Interstate 80 Freeway Dataset (I-80) and the US Highway 101 Dataset (US 101). In both cases, traffic information of a specific road segment was recorded for a period of time. The available information contains trajectories of every recorded vehicle and additional information, like lane boundaries. There are 6 highway lanes available, as well as an off- and on-ramp. See Fig. \ref{figureus101} for a visualization. Measurements were updated 10 times a second.

\subsection{Action-Based Labeling}
To the best of our knowledge, all previous works use a principle which could be called ``action-based labeling'' to generate meaningful labels for data frames \cite{7795631, BALAL201647, 7576883}. The idea is to look for actual occurring lane changes and this way learn from the behavior of the human drivers. A lane change maneuver begins once a vehicle moves towards the lane boundary without oscillations and with a certain speed (0.213 m/s), and terminates when the vehicle crosses the lane boundary, in accordance with the definition in Nie et al. \cite{7795631}. This period of time is called $T_P$, and all included data frames are labeled with $l^ t = 1$. Unfortunately, negative samples are not as well defined or easily spottable, as the drivers' interior motives are not observable: existing works thus assume that before the execution of a lane change the driver takes a certain period of time $T_N$ to analyze the situation, assessing it as not suited for the lane change until eventually a suited situation occurs. All points in $T_N$ are labeled  with $l^t = 0$. 

\begin{figure}[!t]
      \centering
      \includegraphics[scale=0.45]{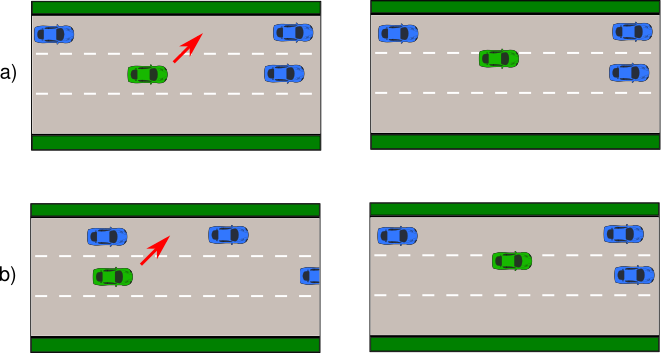}
      \caption{In both scenarios the lane change of the target car (displayed in green) happens in the second frame. In a) the situation several seconds before the lane change, i.e. in $T_N$, looks very similar, thus we deem this example unsuited. In scenario b) the situation changed significantly, thus this example has a higher chance of having a useful label.}
      \label{figuresituation}
\end{figure}

We extend this approach by filtering out examples which present no information gain for any machine learning algorithm.
Intuitively, we only want to include situations in our learning set, in which there was a significant change from the negatively labeled data frame to the positively labeled one, and the situation changed from probably less suited for a lane change to probably more suited (see Fig. \ref{figuresituation}). Additionally, we require a minimum time gap of 1 second between samples with different labels (which translates to a not weighted ``maybe'' class for the LSTM networks).

This gives the following formal definition: Let $t_n \in T_N$ and $t_p \in T_P$. Then 
\begin{equation}
\begin{split}
ad &= d_{PV} + \gamma * d_{PLV} + d_{RV} + \gamma * d_{PFV}  \\
&+ \beta * (\vert v_{PV} \vert + \gamma * \vert v_{PLV} \vert + \vert v_{RV} \vert + \gamma * \vert v_{PFV} \vert) \\
sd &= v_{PV} + \gamma * v_{PLV} - v_{RV} - \gamma * v_{PFV} - ad
\end{split}
\end{equation}
with $\gamma = 2$ and $\beta = 1.8$. These values were chosen to represent the relative importance of the two lanes, as well as average driving behavior on highways.
The sample pair $(t_n, t_p)$ is only included in the learning set if $\frac{\vert ad_{t_n} - ad_{t_p} \vert}{ad_{t_n}} \ge 0.35 $ (1) and $sd_{t_n} \ge sd_{t_p}$ (2). (1) ensures enough relative change in situations, (2) the better suitability of the positive example. That $sd_t$ is indeed an approximation of the degree of suitability can easily be seen by inserting different variable values. 
We call the NGSIM datasets annotated with this method ``action-based dataset''.

When using recurrent models one concern of this approach is a possible overfitting of the networks to the certain temporal structure exhibited by the created sequences, always starting with negative labels and ending with positive ones. To prevent this we augment our data to make the distribution look more random and diverse by adding prolonged sequences, starting and ending at random time points and occasionally containing only positive or negative labels.
{
\setlength\extrarowheight{2pt}
\begin{table}[b!]
\caption{Results on both datasets. A denotes the action-based dataset, B the automatic.}
\label{table_results}
\begin{center}
\tabcolsep=0.15cm
\begin{tabular}{| c | c | c | c | c | c | c |}
\hline
 & IDM & SVM & SVM$^*$ & LSTM & Bi-LSTM$^*$ & Bi-LSTM\\
\hline
A & - & 77.24\% & 78.62\% & 88.76\% & 92.59\% & 88.19\% \\
\hline
B & 61.10\% & 80.70\% & 57.90\% & 83.08\% & 88.49\% & 87.03\% \\
\hline
\end{tabular}
\end{center}
\end{table}
}
\subsection{Automatic Labeling}
As alternative to the method described in the previous subsection we propose an automatic labeling scheme. For this, we run twice over the entire data set, once setting the left lane as hypothetical target lane, once the right lane. For each data frame $t$ we examine all data frames $f^i$ up to 3 seconds in the future and evaluate the observed time gaps in the target lane ($t_{PFV}$ and $t_{PLV}$). If for each examined data frame $f^i$ $t_{PFV} \ge 1$ and $t_{PLV} \ge 1$, $l^t = 1$, otherwise $l^t = 0$. The idea behind this labeling method is that a situation is deemed suitable for a lane change, if a lane change can be executed while all participants can continue driving close to their original speed, i.e. no hard acceleration or braking is necessary. 
Note that this method cannot be used as a simple and perfect (with respect to this labeling) situation assessment algorithm, as it uses information about future states, which are not available in an online application.
We call the NGSIM datasets annotated with this scheme ``automatic dataset''.

When comparing labeled data frames from the action-based labeling to the labeling given by the automatic labeling, about 75\% of the labels match.
Advantages of the automatic labeling method are the number of labeled samples, which now is the full data set, compared to the fraction of data frames around lane change events in the action-based labeling scheme. Further, we prevent the problem of wrong negative labels due to the inobservability of the drivers' intentions and are able to manually model more aggressive or passive drivers by changing the minimum time gaps. 
On the downside, the labels are created by a theoretical concept rather than by actual human behavior. Further, in dense traffic situations fixed time gaps are sometimes not applicable in practice - in order to manage a lane change, a driver might have to aggressively push into a gap and expect other drivers to react.


\section{RESULTS}
\label{sectionresults}
In this section we present our findings from evaluating the different algorithms. First we briefly describe an SVM implementation for a competitive evaluation of our novel algorithm.
We test all models on both the action-based and the automatic dataset, and also extend the SVM approach with future trajectory data to obtain a fair comparison.
For the predictive approaches we test with real future data as a theoretical upper bound for performance, as well as use the IDM for a realizable prediction to show that this bound can (almost) be reached.

\subsection{Support Vector Machine Model}
A reimplementation of the SVM model presented by Nie et al. \cite{7795631} is used. The examples are chosen accordingly by random, but balanced, sampling from all labeled data frames. The approach is extended by equipping each sample with future data frames (5 and 10 seconds later), in order to examine the influence of the prediction component.
\subsection{Evaluation}
The algorithms under examination are the standard SVM approach (SVM), the SVM approach making use of real future trajectories (SVM$^*$), the presented LSTM network (LSTM), the Bidirectional LSTM network with real future trajectories (Bi-LSTM$^*$) and the Bidirectional LSTM with trajectories predicted from the IDM (Bi-LSTM). For the automatic dataset, also a baseline using only the IDM is given: For this the prediction is the resulting automatic label while using the IDM to estimate future vehicle positions in the following 3 seconds.
For evaluation both the datasets I-80 and US 101 were merged and used.  5-fold cross-validation was used for the LSTM networks. For the SVM models the average results of 5 runs with an 80:20 ratio of training and test examples are shown,  since the execution of the SVM approaches involves random sampling from the dataset.
During the splitting into training and test set every track and lane change maneuver was always fully assigned to either the training or the test set, never partially to both sets.
For the SVM models, a Gaussian radial basis function worked best. A grid search was performed to find the best values for the relevant parameters. Different experiments led to the used LSTM parameters. A single layer was used consisting of 128 hidden units, the regularization parameter was set to 0.001.
\begin{figure}[!t]
      \centering
      \includegraphics[scale=0.27]{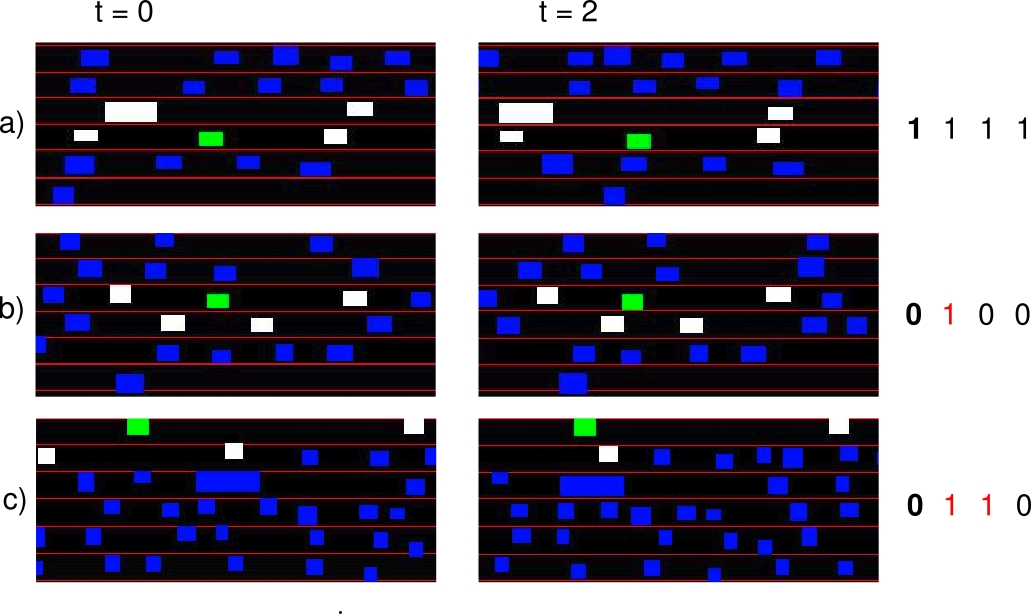}
      \caption{Visualization of different scenarios from the automatic dataset. The ego car is drawn in green, the important surrounding vehicles in white, which indicates the target lane. The situation fed to the algorithms is the one at $t=0$. For each scenario the development after 2 seconds is shown. Next to the scenarios the correct label and the predicted labels from the SVM, the LSTM and the Bidirectional LSTM are shown in this order. A ``1'' defines a situation which is suitable for a lane change, a ``0'' one that is not.}
      \label{figureresult1}
   \end{figure} 
   
\begin{figure}[!t]
      \centering
      \includegraphics[scale=0.4]{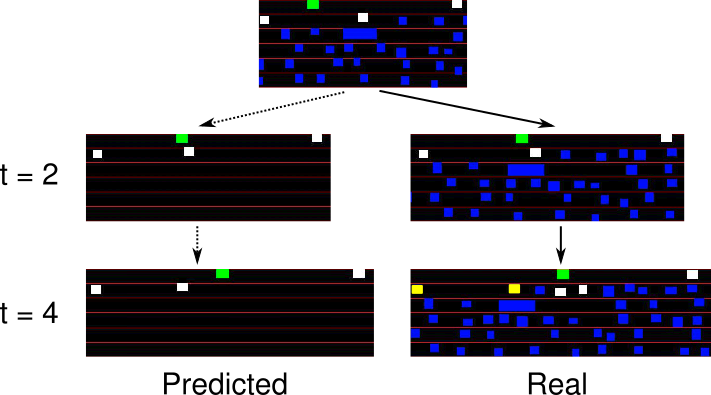}
      \caption{A closer analysis of situation c) from Fig. \ref{figureresult1}. The situation is drawn 2 and 4 seconds later, once as predicted by the IDM and once as actually unfolded.}
      \label{figureresult2}
   \end{figure} 
The results are shown in Table \ref{table_results}.   
As can be seen, already the first LSTM approach outperforms the SVM methods on both datasets significantly. For the SVM approach only the real future trajectories were passed as inputs and the method was not tried in combination with the IDM, since already this theoretical perfect prediction model does not produce better results than the LSTM approaches. We also see, that the Bidirectional LSTM with real future trajectories further improves the classification score on both datasets. When substituting these with predictions from the IDM a small deterioration in accuracy is observed, which is to be expected. For the action-based dataset the results are comparable to those of the non-predictive LSTM network, indicating a relative accurate prediction. For the automatic dataset using the Bidirectional LSTM with predicted trajectories proves to be an improvement of the non-predictive case.
Possible explanations are the better suitability of the long continuously labeled tracks of the automatic dataset, and the lack of questionable labels compared to the action-based dataset.
The IDM baseline alone performs poorly, proving that the combination of situation assessment with deep learning and prediction is needed. 

\begin{figure*}[!t]
    \centering
  \includegraphics[width=\textwidth]{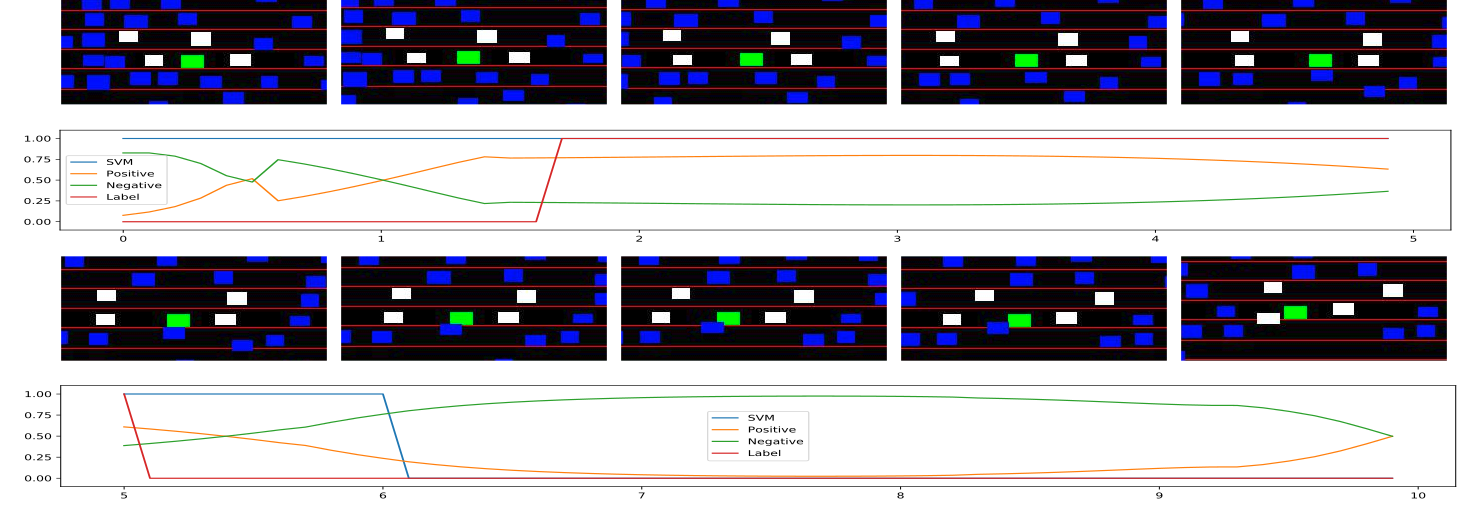}
   \caption{Temporal development of a situation. The images are each taken 1 second apart. In the diagram below the correct label of each frame is displayed as well as the prediction from the SVM and the Bidirectional LSTM. Note that for this the probability $p^t$ for $o^t=1$ is shown (the output of the softmax layer) - compared to the discrete outputs 0 and 1 for SVM and ground truth - meaning the predicted output is $1$ if $p^t > 0.5$.}
   \label{fullres}
\end{figure*}

Fig. \ref{figureresult1} shows three scenarios from the automatic dataset.  Scenario a) is recognized by the SVM, the LSTM and the Bidirectional LSTM (using predicted trajectories) approach correctly as suited for a lane change. In scenario b) the SVM fails, as it does not weigh the velocity of the approaching car in the target lane enough. In scenario c) also the LSTM approach fails. While the situation looks safe at $t=0$, the ego vehicle will have approached the preceding car in the target lane 2 seconds later. This is accurately predicted by the IDM, as can be seen in Fig. \ref{figureresult2}. Note that at $t=4$ EGO will have overtaken PFV, which is not considered in our prediction model: indeed, including these dynamics in our model represents an interesting future research direction.
Fig. \ref{fullres} shows the temporal development of an exemplary situation over 10 seconds. The scene starts in congested traffic, the situation is unsuited for a lane change. As the cars move, a suitable gap forms around the ego vehicle, which is anticipated by the Bidirectional LSTM, although a bit too early. Eventually the gap closes again due to a fast approaching car from behind, which is assessed relatively accurately by the Bidirectional LSTM. The SVM performs worse, it is less accurate and does not handle the temporal changes well. Note that a classification accuracy of 88\% does not mean that 12\% of the sequences are completely mistaken, but instead that of all frames 12\% are misclassified (see Fig. \ref{fullres}: the prediction output of the Bi-LSTM closely matches the ground truth, except the exact moments of label change do not align perfectly). By using for example a delayed or secure action planning, or an additional filter over multiple frames on top, one can expect very safe driving behavior.
\section{CONCLUSIONS AND FUTURE WORK}
In this paper we considered the problem of assessing situations with respect to their suitability for a lane change maneuver. We proposed a novel deep learning architecture based on LSTM networks. We also introduced a novel labeling technique and discussed drawbacks and limitations of the approach used by previous works. We tested our method on the publicly available NGSIM dataset and achieved an average accuracy of 83.08\% - 88.76\%, outperforming existing algorithms. By extending our model to a Bidirectional LSTM with an integrated prediction component, for which we used the IDM, our accuracy improved to 87.03\% - 88.19\%.
Possible applications are the integration of our system into ADAS or fully self-driving cars, supporting these in making decisions when to safely execute a lane change. 
We believe our architecture is applicable to a wide range of problems by simply switching the underlying dataset, giving a method to use supervised learning with recorded driving behavior to assess the feasibility of many different maneuvers. 
In the future we would like to extend and apply our method to these different situations, e.g. include rural and urban scenarios for lane changes and test different maneuvers like yielding and merging at crossroads and roundabouts.
Additionally, we would like to create a dataset labeled by humans according to their situation assessment and compare this to the two labeling methods used here.
Another interesting point of work is the improvement of the prediction component, here instead of the IDM many different methods, amongst others LSTMs, are thinkable of, possibly improving accuracy and narrowing the gap between realizable prediction and the upper bound of a perfect prediction.
Further, integrating additional sensor readings could improve performance of the LSTM model, which could easily be done by filling the occupancy grid with all detected objects.



{\small
	\bibliographystyle{IEEEtran}
	\bibliography{bibliography}

\begin{thebibliography}{10}
\providecommand{\url}[1]{#1}
\csname url@samestyle\endcsname
\providecommand{\newblock}{\relax}
\providecommand{\bibinfo}[2]{#2}
\providecommand{\BIBentrySTDinterwordspacing}{\spaceskip=0pt\relax}
\providecommand{\BIBentryALTinterwordstretchfactor}{4}
\providecommand{\BIBentryALTinterwordspacing}{\spaceskip=\fontdimen2\font plus
\BIBentryALTinterwordstretchfactor\fontdimen3\font minus
  \fontdimen4\font\relax}
\providecommand{\BIBforeignlanguage}[2]{{%
\expandafter\ifx\csname l@#1\endcsname\relax
\typeout{** WARNING: IEEEtran.bst: No hyphenation pattern has been}%
\typeout{** loaded for the language `#1'. Using the pattern for}%
\typeout{** the default language instead.}%
\else
\language=\csname l@#1\endcsname
\fi
#2}}
\providecommand{\BIBdecl}{\relax}
\BIBdecl

\bibitem{ZHENG201416}
Z.~Zheng, ``Recent developments and research needs in modeling lane changing,''
  \emph{Transportation Research Part B: Methodological}, 2014.

\bibitem{8005678}
N.~Motamedidehkordi, S.~Amini, S.~Hoffmann, F.~Busch, and M.~R. Fitriyanti,
  ``Modeling tactical lane-change behavior for automated vehicles: A supervised
  machine learning approach,'' in \emph{Int. Conf. on Models and Technologies
  for Intelligent Transportation Systems (MT-ITS)}, 2017.

\bibitem{trafficsafetly}
U.~D. of~Transportation, ``Critical reasons for crashes investigated in the
  national motor vehicle crash causation survey - february 2015,'' 2015.

\bibitem{tsa}
T.~S.~A. National~Highway, ``Fars encyclopaedia - vehicles involved in single-
  and two-vehicle fatal crashes by vehicle manoeuvre,'' 2009.

\bibitem{NIPS2007_3293}
U.~Syed and R.~E. Schapire, ``A game-theoretic approach to apprenticeship
  learning,'' in \emph{Advances in Neural Information Processing Systems
  (NIPS)}, 2008.

\bibitem{pmlr-v9-ross10a}
S.~Ross and D.~Bagnell, ``Efficient reductions for imitation learning,'' in
  \emph{Int. Conf. on Artificial Intelligence and Statistics (AISTATS)}, 2010.

\bibitem{DBLP:journals/corr/MnihKSGAWR13}
V.~Mnih, K.~Kavukcuoglu, D.~Silver, A.~Graves, I.~Antonoglou, D.~Wierstra, and
  M.~Riedmiller, ``Playing atari with deep reinforcement learning,''
  \emph{arXiv preprint arXiv:1312.5602}, 2013.

\bibitem{7795631}
J.~Nie, J.~Zhang, X.~Wan, W.~Ding, and B.~Ran, ``Modeling of decision-making
  behavior for discretionary lane-changing execution,'' in \emph{Int. Conf. on
  Intelligent Transportation Systems (ITSC)}, 2016.

\bibitem{BALAL201647}
``A binary decision model for discretionary lane changing move based on fuzzy
  inference system,'' \emph{Transportation Research Part C: Emerging
  Technologies}, 2016.

\bibitem{7313255}
S.~Ulbrich and M.~Maurer, ``Situation assessment in tactical lane change
  behavior planning for automated vehicles,'' in \emph{Int. Conf. on
  Intelligent Transportation Systems (ITSC)}, 2015.

\bibitem{7576883}
Y.~Dou, F.~Yan, and D.~Feng, ``Lane changing prediction at highway lane drops
  using support vector machine and artificial neural network classifiers,'' in
  \emph{Int. Conf. on Advanced Intelligent Mechatronics (AIM)}, 2016.

\bibitem{bmw21}
D.~Muoio, ``Bmw plans to take on mercedes by releasing a fully driverless car
  by 2021,'' \emph{Business Insider}, 2017.

\bibitem{tesla21}
F.~Lambert, ``Elon musk clarifies tesla’s plan for level 5 fully autonomous
  driving: 2 years away from sleeping in the car,'' \emph{electrek}, 2017.

\bibitem{A1}
M.~Mukadam, A.~Cosgun, A.~Nakhaei, and K.~Fujimura, ``Tactical decision making
  for lane changing with deep reinforcement learning,'' in \emph{NIPS Workshop
  on Machine Learning for Intelligent Transportation Systems (MLITS)}, 2017.

\bibitem{nagel}
K.~{Nagel}, D.~E. {Wolf}, P.~{Wagner}, and P.~{Simon}, ``Two-lane traffic rules
  for cellular automata: A systematic approach,'' 1998.

\bibitem{6200871}
M.~Ardelt, C.~Coester, and N.~Kaempchen, ``Highly automated driving on freeways
  in real traffic using a probabilistic framework,'' \emph{Transactions on
  Intelligent Transportation Systems}, 2012.

\bibitem{7995915}
C.~Men\'{e}ndez-Romero, F.~Winkler, C.~Dornhege, and W.~Burgard, ``Maneuver
  planning for highly automated vehicles,'' in \emph{Intelligent Vehicles
  Symposium (IV)}, 2017.

\bibitem{7565491}
J.~Morton, T.~A. Wheeler, and M.~J. Kochenderfer, ``Analysis of recurrent
  neural networks for probabilistic modeling of driver behavior,''
  \emph{Transactions on Intelligent Transportation Systems}, 2017.

\bibitem{7995938}
S.~G. Jeong, J.~Kim, S.~Kim, and J.~Min, ``End-to-end learning of image based
  lane-change decision,'' in \emph{Intelligent Vehicles Symposium (IV)}, 2017.

\bibitem{2000PhRvE..62.1805T}
M.~Treiber, A.~Hennecke, and D.~Helbing, ``Congested traffic states in
  empirical observations and microscopic simulations,'' \emph{Phys. Rev. E},
  2000.

\bibitem{gers1999learning}
F.~A. Gers, J.~Schmidhuber, and F.~Cummins, ``Learning to forget: Continual
  prediction with lstm,'' 1999.

\bibitem{DBLP:journals/corr/MikolovSCCD13}
T.~Mikolov, I.~Sutskever, K.~Chen, G.~S. Corrado, and J.~Dean, ``Distributed
  representations of words and phrases and their compositionality,'' in
  \emph{Advances in Neural Information Processing Systems (NIPS)}, 2013.

\bibitem{ngsim}
``Ngsim project,'' https://ops.fhwa.dot.gov/trafficanalysistools/ngsim.htm.

\end{thebibliography}
}
\end{document}